# Primender Sequence: A Novel Mathematical Construct for Testing Symbolic Inference and AI Reasoning

Mohd Anwar Jamal Faiz

Senior Member, IEEE
Independent Researcher, India

Email: Toughjamy@yahoo.com

*Abstract* - **This paper introduces the Primender sequence, a novel integer sequence defined by a hybrid rule that combines classical primality with modular digit-based conditions. Specifically, a number n is included in the sequence if it is prime or ends with a prime number of unit digit or any length. In other words, numbers which are primes or have at least one prime suffix. The resulting sequence exhibits a deterministic yet non-trivial structure, blending number-theoretic properties with symbolic patterning. We propose the Primender sequence as a benchmark for evaluating the symbolic reasoning capabilities of Large Language Models (LLMs). The study is motivated by the need for interpretable, rule-based testbeds that can assess an LLM's ability to infer hidden rules, validate mathematical hypotheses, and generalize symbolic logic at scale. A key hypothesis explored is: Whenever a number in the Primender sequence is exactly one more than the largest prime less than or equal to it, the difference between it and the previous number in the sequence is also 1. We design a structured prompt and evaluation framework to test this hypothesis across multiple state-of-the-art LLMs, including ChatGPT, Copilot, DeepSeek, Gemini, Grok, and LLaMA. The models are tasked with identifying the underlying rule, validating the hypothesis, and generating the next 100,000 terms of the sequence. Comparative metrics such as rule inference accuracy, hypothesis evaluation, sequence validity, and symbolic explanation quality are used to assess model performance. This work contributes a novel mathematical construct and a reproducible methodology for benchmarking LLMs in symbolic reasoning, hypothesis testing, and scalable pattern generalization - bridging the domains of number theory, artificial intelligence, and software engineering.**

## I. INTRODUCTION

Prime numbers have long fascinated mathematicians due to their fundamental role in number theory and their unpredictable distribution. Defined as natural numbers greater than 1 that have no positive divisors other than 1 and themselves, primes are the building blocks of the integers [1][2]. Their applications span from pure mathematics to modern cryptography, random number generation, and algorithm design [3]. In this paper, we introduce a novel integer sequence called the Primender sequence.

The name Primender is a portmanteau of "prime" and "ender", reflecting the sequence's defining characteristic: it includes numbers that are either prime or end with a digit combination that is a prime number. More formally, a number n is included in the Primender sequence if it satisfies any of the following conditions:

    n is a prime number, or

    n mod     $10 \in \{2, 3, 5, 7\}$, or

    n mod     100 is a prime number, or

    n mod     1000 is a prime number, and so on.

This hybrid rule blends classical primality with digit-based modular conditions, resulting in a deterministic yet non-trivial sequence that exhibits both mathematical structure and symbolic complexity. The motivation behind constructing the Primender sequence is twofold. First, it offers a new lens through which to explore the interplay between digit patterns and primality. Second, and more significantly, it serves as a symbolic benchmark for evaluating the reasoning capabilities of large language models (LLMs). As LLMs increasingly assist in mathematical problem-solving [4] and symbolic inference, it becomes essential to assess their ability to identify hidden rules, validate hypotheses, and generalize patterns at scale.

To this end, we formulate a hypothesis based on derived properties of the sequence: Whenever a number in the sequence is exactly one more than the largest prime less than or equal to it (i.e., $PE_n - LP_n = 1$), the difference between it and the previous number in the sequence is also 1. This hypothesis is used to test the symbolic reasoning and hypothesis evaluation capabilities of various LLMs. Furthermore, we challenge these models to extend the sequence by generating the next 100,000 terms based solely on the initial 100 terms - without being explicitly told the rule. This paper presents the formulation of the Primender sequence, its mathematical properties, and a comprehensive evaluation framework for testing LLMs using this sequence. The results offer insights into the symbolic reasoning potential of modern AI systems and propose a novel benchmark for future research in explainable and interpretable AI.

## II. RESEARCH METHODOLOGY

Our research methodology introduces a structured, symbolic approach to evaluate the reasoning, pattern recognition, and explainability capabilities of large language models (LLMs). This study employs a Symbolic Constructivist Evaluation Methodology (SCEM), combining exploratory sequence design, hypothesis-driven symbolic pattern discovery, and prompt-based evaluation of large language models. The Primender sequence serves as a symbolic benchmark [5] to assess LLMs' capabilities in rule induction, explainability, and decision support.

It consists of three core stages:

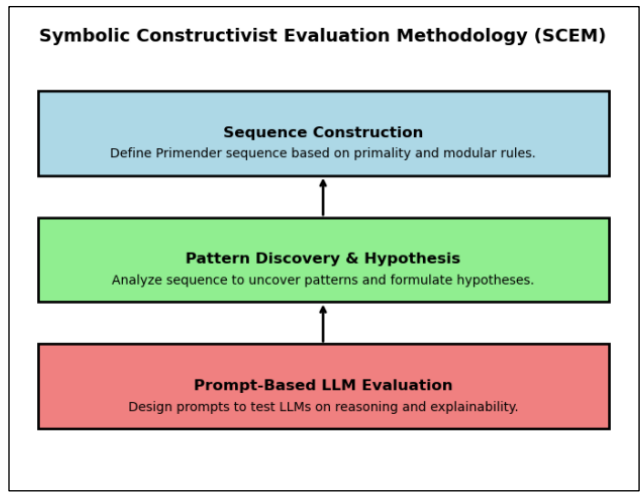

Fig 1 - the SCEM approach used in the research

*A. Devising an original Primender Series*

We are introducing a novel number sequence Primender and exploring its mathematical and symbolic properties. This method is often used in early-stage AI research to define new constructs or frameworks, especially in symbolic AI and explainability.

*B. Discover a hidden pattern and propose a valid hypothesis*

We are constructing a new symbolic benchmark (Primender sequence) and using it to evaluate LLMs. This is common in computer science when proposing new algorithms, models, or symbolic systems.

*C. Constructing a Prompt and use it to test LLMs*

We are testing LLMs using structured prompts and measuring their performance on symbolic reasoning tasks. This is standard in AI benchmarking [6], especially in explainable AI (XAI) and decision support systems.

### III. PRIME NUMBERS AND THE PRIMENDER SEQUENCE

Prime numbers are natural numbers greater than 1 that have no positive divisors other than 1 and themselves. They are the fundamental building blocks of number theory, much like atoms in chemistry. Every integer can be uniquely factored into primes, a principle known as the Fundamental Theorem of Arithmetic [7]. Primes are crucial in modern applications such as cryptography, computer algorithms, and digital security systems. They also play a vital role in hashing, random number generation, and error detection. Their simplicity and indivisibility make them indispensable in both theoretical mathematics and practical computing systems.

Prime numbers have fascinated mathematicians for centuries due to their unpredictable distribution and elusive patterns. Despite their simple definition, no formula can generate all primes, and their gaps grow irregularly. The mystery deepens with unsolved problems like the Riemann Hypothesis and Goldbach's Conjecture [8][[9]. Their randomness within order challenges our understanding of number theory. This paradox - being both fundamental and enigmatic - has made primes a timeless subject of intrigue, inspiring generations of mathematicians to explore their secrets.

Prime numbers form various intriguing subsets, each with unique properties and applications. Examples include twin primes (pairs like 11 and 13), Mersenne primes (of the form $2^n-1$), Sophie Germain primes, and palindromic primes. These subsets are studied for their roles in cryptography, primality testing, and mathematical proofs. Mersenne primes, for instance, are used in generating large prime keys for encryption. Studying these subsets helps uncover deeper patterns in number theory and enhances our understanding of computational complexity and algorithmic efficiency.

Let's start with the Primender Sequence rules recap. A number x belongs to Primender Series if any of the following is true:

x is a prime number, or

x mod 10 ∈ { 2 , 3 , 5 , 7 }, or

x mod 100 is a prime number, or

x mod 1000 is a prime number, and so on.

Now let's discuss for the Primender series (numbers that end with primes or whose endings match prime conditions), and few notable characteristics that make it mathematically interesting and visually appealing.

*A. Digit-Based Prime Signature*

The Primender series is defined not by traditional primality, but by a positional relationship to primes — a number qualifies if its last 1, 2, or 3 digits form a prime number. This creates a fascinating hybrid between number theory and digit patterns, opening up creative visualizations (e.g., trees of numbers growing from prime roots).

We used Python 3.13 and matplotlib library to create charts and do the computations for this research paper. Here's a detailed analysis of the first 10000 numbers that satisfy the selection rules, along with their delta values (differences between consecutive terms).

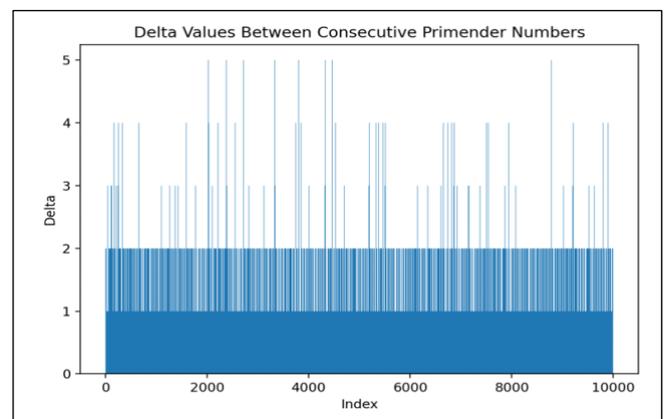

*Fig 2 - the pattern how frequently small or large jumps occur, which can be useful for analyzing symbolic regularity and non-linear growth in AI pattern recognition tasks.*

When extended, here's a detailed analysis of the first 100000 numbers that satisfy the selection rules, along with their delta values.

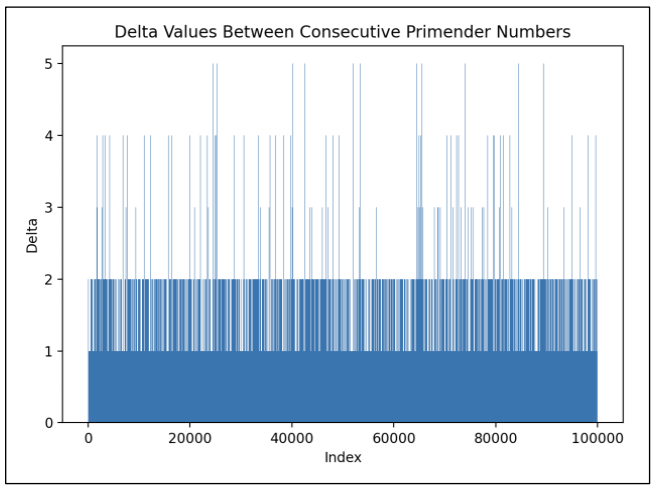

*Fig 3 - analysis of the first 100000 numbers*

The above chart also shows a similar pattern - lower delta values are more frequent than the higher ones. The Python code is given below and is also available on Github repository for Primender [10].

```
import matplotlib.pyplot as plt
# Function to check if a number is prime
def is_prime(n):
    if n <= 1:
        return False
    if n <= 3:
        return True
    if n % 2 == 0 or n % 3 == 0:
        return False
    i = 5
    while i * i <= n:
        if n % i == 0 or n % (i + 2) == 0:
            return False
        i += 6
    return True

# Function to check if a number is a Primender
def is_primender(n):
    return is_prime(n) or n % 10 in {2, 3, 5, 7} or is_prime(n % 100) or is_prime(n % 1000)
# Generate the first 10,000 Primender numbers
primender_numbers = []
n = 1
while len(primender_numbers) < 100000:
    if is_primender(n):
        primender_numbers.append(n)
    n += 1

# Compute delta values
deltas = [primender_numbers[i] - primender_numbers[i - 1] for i in range(1, len(primender_numbers))]

# Plot the bar chart
plt.bar(range(1, len(deltas) + 1), deltas, width=1.0)
plt.xlabel('Index')
plt.ylabel('Delta')
plt.title('Delta Values Between Consecutive Primender Numbers')
plt.tight_layout()
plt.show()
```

## B. Maximum Delta Value is 5

- An interesting pattern that we can note is that Delta does not ever crosses the value 5. So, this means that the maximum value of the delta is 5. We saw this even when we ran our program for first 1 million Primender numbers. This is contrast with Prime gaps [11].
- Histogram of Delta Frequencies reveals how frequently each delta value occurs. It highlights the most common gaps between consecutive Primender numbers, which is useful for understanding local density and regularity. Python code are available online in the folder *Sample-Python-Scripts-For-Primender-Numbers* of the Primender repository [10].

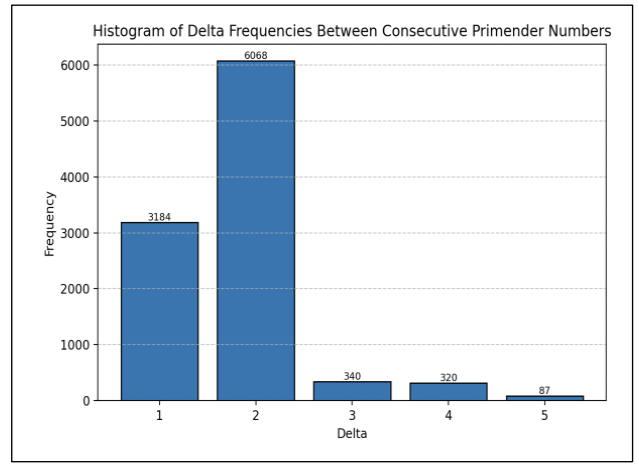

*Fig 4 - visualization that helps identify the most common gaps and the spread of spacing in the sequence for first 10000 Primenders.*

- Delta vs. Primender Index (Scatter Plot [12]) shows how delta values vary across the sequence. The python code for this is available online in Primender repository [10].

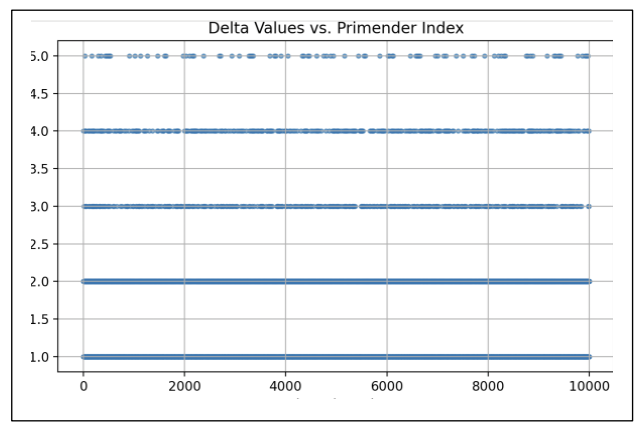

*Fig 5 – Scatter plot to detect patterns, bursts, or anomalies in the spacing of Primender numbers.*

## C. Mathematical proof that maximum delta possible is 5

Assume a delta Δ > 5 exists between two consecutive Primender numbers. Then, all integers in that gap must fail all four inclusion rules. However, within any 6 consecutive integers, at least one will:

- Be prime (by Bertrand's postulate [13] for small gaps),
- Or end in 2, 3, 5, or 7,
- Or have a last two, three (or four etc.) digits that are prime.

Thus, it is impossible to construct a gap of more than 5 without violating the inclusion rules. Therefore, the maximum delta in the Primender sequence is 5. More details and a beautiful mathematical proof by induction can be found on the blog introducing Primender sequence online [14].

*D. Some notable observations in Primender sequence*

- The Primender sequence is densely packed, with most numbers being close to each other. The rules favor small increments, especially due to the inclusion of numbers ending in 2, 3, 5, or 7.

- The presence of primes and modular primes (last 2 or 3 digits) ensures a steady stream of qualifying numbers. Hence, the Primender sequence is a never ending series and will have infinite numbers.

- As discussed, there are no instances where the delta (difference between consecutive numbers) is exactly 6 or more. This suggests that the selection criteria produce a sequence where numbers are relatively close together, and a jump of 6 is too large to occur under the given rules.

- For the sake of the research, we have compiled all the 1 million data points in table format. The Excel file *million-primender-numbers-with-deltas* is published online [10] contains the first 1,000,000 numbers that satisfy the selection rules, along with their corresponding delta values.

- Every time the delta is 4, the second number in the pair of Primender indeed ends with the digit 1. Examples of such pairs: (7, 11) (37, 41) (67, 71) (97, 101). This hypothesis was confirmed with the first 100,000 selected numbers. There are 1,069 instances where the delta is 4, and in every case, the second number in the pair ends with the digit 1.

- We have also analyzed within Primender series to find index of those numbers which are actually prime in completeness. Out of first 1000 numbers, actual primes (i.e., numbers that are prime in full, not just by ending) appear at the following indices (0-based) eg. 0, 1, 2, 3, 4, 6, 8, 9, 11, 14, 15, 19, and so on. On full analysis we saw that the primes are evenly scattered throughout the sequence and there are no large gaps or long stretches without primes. This suggests that true primes remain consistently interspersed throughout the sequence.

The Primender sequence opens a rich frontier for mathematical and computational exploration. Its deterministic yet non-trivial structure offers a symbolic framework for studying prime distribution, modular arithmetic, and pattern emergence. As this is a novel and original number sequence, never discussed in literature before, we have also added it to the Online Encyclopedia of Integer Sequences (OEIS database).

This newly catalogued OEIS sequence A384735 [15] invites researchers to investigate its potential in cryptographic key generation, pseudo-random number design, and symbolic AI reasoning. Its structured unpredictability makes it a compelling candidate for training interpretable machine learning models and benchmarking generative AI systems. We have also created a tool Primender Checker [16] to enable students, researchers and amateur mathematicians to check if a number is Primender or not and find its index and other properties. We encourage scholars to cite and explore the Primender series by the author of this research as a foundational tool in symbolic computation and intelligent systems.

## IV. DESIGNING A HYPOTHESIS

A mathematical hypothesis is a proposed explanation or conjecture based on observed patterns, logical reasoning, or theoretical insight. It serves as a starting point for investigation, aiming to be tested, proven, or disproven through formal mathematical methods, such as deduction, counterexamples, or computational verification.

To design a new hypothesis for the Primender sequence, we followed a rigorous pattern discovery methodology rooted in symbolic reasoning. We had to identify some pattern or anomaly and first verify its truthfulness ourself i.e. frame it precisely to allow mathematical proof or falsification.

We began by compiling the first 1 million Primender numbers into an Excel dataset *million-primender-numbers-with-deltas* [10]. For each entry, we introduced the following key data points:

- $PE_n$: The nth Primender number (e.g., 2, 3, 5, 7, 11, 12, 13, …)
- $LP_n$: The largest prime number less than or equal to $PE_n$
- Delta: The difference between consecutive Primender numbers.

By visually inspecting and analyzing this structured data, we observed a recurring micro-pattern: whenever a Primender number was exactly one more than the largest prime less than or equal to it (i.e., $PE_n - LP_n = 1$), the delta from the previous Primender was also 1. This led us to formulate the following hypothesis:

If $PE_n - LP_n = 1$, then $\Delta = 1$. We then verified this hypothesis computationally for the all first 1000 Primender numbers, and found it to hold true without exception. Later we verified it on all the Primender numbers.

Finally, we wrote a Python code to test out our hypothesis for first 1 Lakh Primender numbers and found the hypothesis to be computationally true. The source code is given below and is also on the Primender Github repository [10].

```
# Function to check if a number is prime
def is_prime(n):
    if n <= 1:
        return False
    if n <= 3:
        return True
    if n % 2 == 0 or n % 3 == 0:
        return False
    i = 5
    while i * i <= n:
        if n % i == 0 or n % (i + 2) == 0:
            return False
        i += 6
    return True
```

```
# Function to check if a number is a Primender
def is_primender(n):
    return is_prime(n) or n % 10 in {2, 3, 5, 7} or is_prime(n % 100)
or is_prime(n % 1000) or is_prime(n % 10000) or is_prime(n %
100000)

# Generate the first 100000 Primender numbers
primender_numbers = []
n = 1
while len(primender_numbers) < 100000:
    if is_primender(n):
        primender_numbers.append(n)
    n += 1

# Compute delta values
deltas = [primender_numbers[i] - primender_numbers[i - 1] for i in
range(1, len(primender_numbers))]

# Function to find the largest prime less than or equal to a given
number
def largest_prime_less_than_or_equal_to(n):
    for i in range(n, 1, -1):
        if is_prime(i):
            return i
    return None

# Verify the hypothesis
hypothesis_holds = True
for i in range(1, len(primender_numbers)):
    PEn = primender_numbers[i]
    LPn = largest_prime_less_than_or_equal_to(PEn)
    if PEn - LPn == 1:
        if deltas[i - 1] != 1:
            hypothesis_holds = False
            print(f"Hypothesis fails at PEn = {PEn}, LPn = {LPn}, Delta
= {deltas[i - 1]}")
            break

if hypothesis_holds:
    print("The hypothesis holds for the first 100000 Primender
numbers.")
else:
    print("The hypothesis does not hold for the first 100000 Primender
numbers.")
```

There are some unique characteristics of LPn and the PEn-LPn series discovered above. Some of notable things are listed below:

### A. Analysis of Dot Matrix Timeline for the first 100 Primender numbers

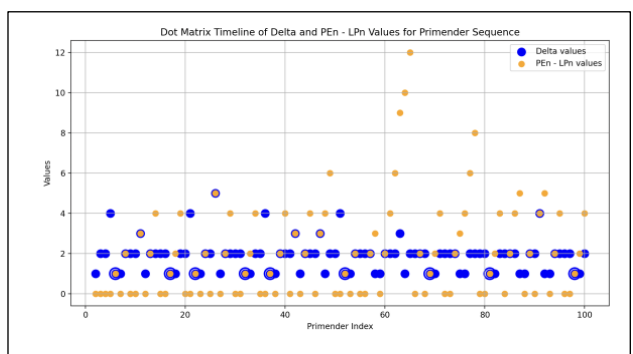

*Fig 6 - Dot Matrix or scatter diagram for two key columns.*

The X-Axis is the Primender index and it represents the position of each number in the Primender sequence (from 1 to 100). The Y-Axis is the Value Magnitude and it represents the numerical values of two metrics:

- Delta: The difference between a Primender number and the one before it.
- $PEn - LPn$: The difference between the Primender number and the largest prime $\leq$ it.

In the chart the Blue dots show the Delta values. These indicate how far apart each Primender number is from the previous one. The Orange dots represent $PEn - LPn$ values. These measure how close each number is to the nearest prime below or equal to it. The Blue Rings around Orange Dots highlights the symbolic coincidences where: Delta = 1 and $PEn - LPn = 1$.

The chart clearly manifests three major points:

1. The delta assumes a maximum value of 5. We have already demonstrated earlier that the Delta value can never be 6 or larger.

2. Whenever PEn-LPn is 1 and an orange dot is marked at value 1, there is a blue dot as well at the same point suggesting that Delta is also 1. That is, an orange dot cannot exist at value 1 in isolation.

3. We see that blue dots may exist at isolation in the dot matrix diagram. This means that Delta can be 1 at multiple PEn-LPn values. That also means that the reverse of our hypothesis is not true.

The full code to reproduce this Dot Line Matrix or Scatter diagram is published online. [10]

### B. PEn-LPn can have any digit at the unit place except 7

When we examined the PEn-LPn values for all the dataset we created we found that they didn't assume the values 7, 17, 27, 37 and so on. The absence of values ending in 7 is not coincidental but a structural artifact of the hybrid rule set. It reflects how symbolic constraints (like digit endings and modular primes) shape the distribution of gaps in the sequence. This feature of any rule based mathematical construct can help study the series in a fresh perspective, even the prime number series itself.

### C. PEn-LPn has no upper limit like Delta had Max value 5

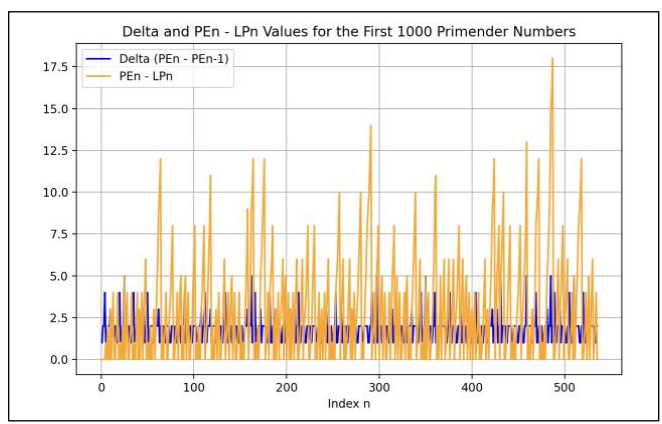

*Fig 7 - line chart comparing two symbolic metrics for the first 1000 Primender numbers*

As numbers grow, prime gaps (the distance between consecutive primes) also grow - a well-known result in number theory [1][2]. By our proposed mathematical construct, not all the Primender numbers are prime - many are included due to symbolic rules (e.g., ends in 3, or mod 100 is 17). If a Primender number is not prime then sometimes the difference with the last actual prime before it could become big. Therefore, $PE_n - LP_n$ can grow without bound, especially when a Primender number is symbolically included far from the last prime.

## V. DESIGN THE PROMPT FOR TESTING

Prompt Design: To simulate a realistic and context-rich interaction, we provided each Large Language Model (LLM) [17] [18] with a single, structured prompt containing the sequence, a derived hypothesis, and a request for rule identification and validation. While the hypothesis may appear to hint at the underlying rule, it does not explicitly define it. The task still requires the model to reason through multiple layers: identifying the rule, computing derived series (LPn and Delta), and evaluating the conditional relationship. This design allows us to assess not only pattern recognition but also symbolic reasoning and hypothesis testing capabilities of the models.

Prompt: *I am working with a custom number sequence that I call the Primender sequence. It is defined by a specific rule.*

*Here are the first 100 terms of the sequence: 2, 3, 5, 7, 11, 12, 13, 15, 17, 19, 22, 23, 25, 27, 29, 31, 32, 33, 35, 37, 41, 42, 43, 45, 47, 52, 53, 55, 57, 59, 61, 62, 63, 65, 67, 71, 72, 73, 75, 77, 79, 82, 83, 85, 87, 89, 92, 93, 95, 97, 101, 102, 103, 105, 107, 109, 111, 112, 113, 115, 117, 119, 122, 123, 125, 127, 129, 131, 132, 133, 135, 137, 139, 141, 142, 143, 145, 147, 149, 151, 152, 153, 155, 157, 159, 161, 162, 163, 165, 167, 171, 172, 173, 175, 177, 179, 181, 182, 183, 185*

Prompt Title: Pattern Discovery, Hypothesis Evaluation, and Sequence Extension Task

*Your task is to:*

1. *Identify the rule that generates this sequence. Explain the logic behind the rule.*

2. *Consider the following two derived series: - LPn: For each number in the sequence, the largest prime less than or equal to it. - Delta: The difference between each number and the previous number in the sequence.*

3. *Evaluate the following hypothesis: > Whenever PEn - LPn = 1, the corresponding Delta is also 1. That is, whenever a number in the sequence is exactly one more than the largest prime less than or equal to it (i.e., PEn - LPn = 1), the difference between it and the previous number in the sequence (Delta) is also 1. Explain whether this hypothesis is true or false, and why.*

4. *Generate the next 100,000 terms of the sequence based on your inferred rule. Ensure that all generated numbers strictly follow the same logic., describe how you verified the correctness of your output.*

*This task is part of a study evaluating how well language models can perform symbolic reasoning, validate hypotheses, and generalize patterns at scale.*

## VI. LLM COMPARISON TABLE FOR PRIMENDER SEQUENCE EVALUATION

In this research, earlier we introduced the Primender Sequence, a novel mathematical construct defined by a hybrid rule combining primality and modular digit conditions. Motivated by the growing need for interpretable, rule-based testbeds, this study proposes the Primender sequence as a tool to assess an LLM's ability to infer hidden rules, validate mathematical hypotheses, and generalize symbolic logic at scale.

We discovered and evaluated a central hypothesis – *'Whenever a number in the sequence is exactly one more than the largest prime less than or equal to it (i.e., $PE_n - LP_n = 1$), the difference between it and the previous number in the sequence (Delta) is also 1.'* - to be true.

Finally, we designed a structured prompt and fed it to different Generative AI chatbots available to evaluate the multiple state-of-the-art LLMs in identification, evaluation and expansion tasks. The three main areas under which the performance was tested are –

*a) Rule inference and accuracy of series logic*

*b) Evaluation of correctness of our proposed hypothesis and the quality of explanation provided, and*

*c) The error rate in the 100000 numbers generated for the series.*

This research bridges number theory, AI, and software engineering, offering a reproducible methodology for benchmarking symbolic intelligence in LLMs.

| LLM Model | Rule Inference Accuracy | Hypothesis Evaluation | Generated Sequence Error Rate |
|---|---|---|---|
| ChatGPT 3.5 | No | Yes | Not generated |
| ChatGPT o3 | Yes | Yes | 5.16% |
| ChatGPT 4 | No | Yes | 19.22% |
| Copilot | No | Yes | 79.43% |
| DeepSeek R1 | No | Yes | 44.90% |
| Gemini 2.5 flash | No | No | Not generated |
| Gemini 2.5 Pro | No | Yes | 99.46% |
| LLaMA 3.3 | No | Yes | Not generated |
| Grok 3 | No | Yes | 79.48% |

*Fig 8 - results of running prompts on different LLMs.*

The evaluation results indicate a diverse range of performance among large language models (LLMs) when tasked with interpreting and reasoning over the Primender sequence framework. While a subset of models demonstrated high accuracy and consistency in reaching the correct conclusions with minimal guidance, others exhibited limitations in symbolic reasoning and required iterative prompting to align with the intended analytical objectives. In several cases, human intervention was necessary to clarify task requirements or reframe prompts to elicit responses in the desired format.

This variability highlights the current disparity in LLMs' capabilities for symbolic inference, rule induction, and structured output generation. The findings underscore the importance of prompt engineering and model interpretability in symbolic evaluation tasks. Moreover, they suggest that while some LLMs are approaching autonomous reasoning proficiency, others still rely heavily on human-in-the-loop strategies to achieve satisfactory performance in complex, rule-based analytical contexts.

A: *Reading final results and Errors in detail*

The tabular result captured above clearly shows our evaluation results using metrics such as rule inference accuracy, hypothesis evaluation, sequence generation capabilities. The Python code we used in comparing the final list of numbers generated by each LLM is given below and is also at Github [10].

```
from collections import Counter

def read_numbers_from_file(filename):
    with open(filename, 'r') as file:
        return [int(line.strip()) for line in file if line.strip().isdigit()]

def compare_unordered_lists(correct_file, test_file):
    correct_seq = read_numbers_from_file(correct_file)
    test_seq = read_numbers_from_file(test_file)
    counter_correct = Counter(correct_seq)
    counter_test = Counter(test_seq)
    # Get all unique numbers from both
    all_keys = set(counter_correct.keys()).union(set(counter_test.keys()))
    matched = 0
    unmatched = 0
    for key in all_keys:
        correct_count = counter_correct.get(key, 0)
        test_count = counter_test.get(key, 0)
        matched += min(correct_count, test_count)
        unmatched += abs(correct_count - test_count)
    total_expected = sum(counter_correct.values())
    total_tested = sum(counter_test.values())
    error_percent = ((total_expected - matched) / total_expected) * 100 if total_expected else 0

    # Optional: breakdown of extra/missing items
    missing = counter_correct - counter_test
    extra = counter_test - counter_correct

    if missing:
        print("\n▼ Missing or less frequent in test file:")
        for num, count in missing.items():
            print(f" • {num} → Missing {count} time(s)")
    if extra:
        print("\n▲ Extra or more frequent in test file:")
        for num, count in extra.items():
            print(f" • {num} → Extra {count} time(s)")

    # Final Summary
    print("\n======= Summary =======")
    print("🔍 Method Used    : Ignore Order – Compare Content")
    print("🧭 Strategy       : Matches frequencies, computes true error")
    print("📊 Metrics        :")
    print(f" • Total Expected    = {total_expected}")
    print(f" • Total Tested      = {total_tested}")
    print(f" • Correct Matches   = {matched}")
    print(f" • Total Errors      = {total_expected - matched}")
    print(f" • Error Percentage  = {error_percent:.2f}%")
    print("✅ Suitable for unordered validation with correct proportions.")

compare_unordered_lists('correct-full-list.txt','result.txt')
```

We found that the Ignore Order Compare Content algorithm is ideal when the position of elements doesn't matter, only their presence and frequency. It treats sequences as multisets and uses tools like *collections.Counter* for efficient comparison. This method is especially useful in validating unordered data, such as survey responses, inventory lists, or generated outputs with flexible ordering. It quickly identifies missing or extra items, supports frequency analysis, and handles duplicates accurately. Its time complexity is linear, making it scalable. The algorithm is deterministic, easy to implement, and provides precise error diagnostics, making it valuable for debugging, quality checks, and content verification tasks.

The observations for each LLM Model are listed below. The full details, python code execution summary, full chat communication and actual response received from each LLM Model is published in the Output folder in the Primender Github Repository [10]

| LLM Model | Notes |
| --- | --- |
| ChatGPT 3.5 | It was quick but little slower than Copilot. However, did not require any second prompt. It evaluated a rule and then identified that it is wrong and then formed another rule. Later stopped more advanced data analysis suggesting data analysis limit reached. Hypothesis evaluation was done with 100 terms. |
| ChatGPT o3 | First it clarifed 3 points and we provided relevant answers. Research completed in 4 minutes. Rule was inferred very closely but since sample were till two digits only, it made rule till last two digits to be prime. Hypothesis evaluation was done with 100 numbers and also an explanation. To download generated number list had to prompt again. |
| ChatGPT 4 | Used 'Think For Longer' Feature of Chatgpt to invoke this model. It took 45 sec to analyze and give results. No human intervention required and downloadable link of all 1 lakh numbers also provided with single prompt. The Hypotheisis was proved to be true theoretically. |
| Copilot | It was quick within 10 seconds and no secondary prompt required. The rule evaluated was wrong. It was similar to gerated by ChatGPT 3.5 version but there was no self feedback loop or alternate rule generation. Hypothesis evaluation was done with 100 terms. |
| DeepSeek R1 | It took good around 3 minutes and gave inference and hypothesis evaluation in single prompt. For list of numbers we asked to be given as downloadable link again. For downloadable file we send a second prompt. It generated a link however the file not downloadable as it was hosted on a server giving errors. Generated list using the program it returned. |
| Gemini 2.5 Flash | Took around 1.5 minute and gave half reply. The pattern found was incorrect and it didn't evaluated the hypothesis. Upon reminding - the hypothesis evaluation was incorrect as it said the our hypothesis was wrong. For list of numbers generation it gave a python code and we gave third prompt to provide downloadable txt file. Finaly it apologized that it can't generate such downloadable file. |
| Gemini 2.5 Pro | Took around 1 minute and gave a rule and hypothesis evaluation in single prompt. For full list of primender sequence as asked earlier, we had to send a prompt again. It still just gave a sample list of primender numbers. On third prompt and reminding that we need downloadable text file, it tells that The full list is too long to display in its entirety but is contained within the copyable block. However, when copied even that had only around 2000 numbers. |

| LlaMA 3.3 | Used DeepInfra tool to access this LLM. Had to provide multiple inputs through prompts. Hypothesis evaluation was correct but when asked to give me list of numbers as txt file it tells that it doesn't have the capability to directly provide a downloadable text file. |
|---|---|
| Grok 3 | The chatbot was quick and response in 17 seconds. The pattern recognition was wrong but hypothesis evaluation was correct. It did theoretically too. For downloadable file as txt had to send prompt again. |

B: *Comparison results correlation of our results:*

Numerous comparative studies have been done for LLMs [19][20] and each identify different areas very each are the most fit. For the research and deeper analysis, you need more dedicated LLMs whose time of execution will be more. On the other hand, faster LLMs are good at providing quick response but on the cost of the depth and quality of the analysis. We evaluated nine state-of-the-art Large Language Models (LLMs) using the Primender sequence benchmark on three key criteria - Rule Inference Accuracy, Hypothesis Evaluation, and Generated Sequence Error Rate.

Only ChatGPT o3 successfully inferred the correct rule and validated the hypothesis with a relatively low error rate (5.16%), making it the most robust model in symbolic reasoning and pattern extrapolation for this task. Most models (8 out of 9) failed at rule inference, highlighting current LLM limitations in inductive symbolic logic despite strong performance in hypothesis evaluation. Gemini 2.5 Pro and Copilot showed high error rates (99.46% and 79.43%, respectively), indicating poor generalization to symbolic sequence generation. Models like ChatGPT 3.5, LLaMA 3.3, and Gemini 2.5 Flash failed to generate a usable sequence, reflecting possible token/response limits or prompt misalignment.

Beyond numerical performance, we also analyzed the usability, interaction flow, and output completeness of each LLM. Key metrics included response time, prompt dependency, interpretability, support for downloadable output, and theoretical reasoning capability. Our results reveal that only ChatGPT o3 and ChatGPT 4 reliably infer or validate symbolic rules in the Primender sequence, with ChatGPT o3 achieving the lowest sequence generation error (5.16%). This aligns with findings in [19], where GPT-4-based models consistently outperform others in symbolic reasoning and multi-step logical deduction. In contrast, Gemini variants, despite being competitive in general NLP tasks in, struggle to generate valid symbolic sequences and correctly evaluate hypotheses in our domain, reflecting a similar limitation noted in the Baeldung AI report's [19] observations on Gemini's weaker rule-based reasoning.

Scale AI's "Humanity's Last Exam" [20] emphasizes the challenge LLMs face in formal reasoning and hypothesis testing, particularly under constrained data or ambiguous prompts. Our study corroborates this: most models, including Copilot and Grok, can often state hypothesis evaluations but fail at consistent sequence extrapolation, highlighting a gap between theoretical reasoning and practical generation.

Our findings also expose real-world limits in large sequence generation at scale (100,000 terms), where models like Gemini and LLaMA fail to provide complete or downloadable outputs. This operational bottleneck is less emphasized in [19] and [20], which often test smaller or token-limited tasks but aligns with recent community reports on LLM scalability challenges in symbolic generation [19].

The iterative prompting required for downloadable output and corrections in our research underlines the observation in humanities last exam [20] that current LLMs often require human-in-the-loop to achieve explainable and reliable outputs in symbolic domains. We also found that Models that asked clarifying questions, such as ChatGPT o3, demonstrated superior rule inference and hypothesis evaluation. This aligns with findings from Self-Ask prompting [21], which shows that LLMs perform significantly better when allowed to query ambiguous aspects of a problem before answering. Such dialogic prompting mirrors human reasoning and enhances symbolic understanding. Our results reinforce that multi-turn engagement and explanatory questioning markedly improve symbolic reasoning accuracy in LLMs.

The prompt-engineering technique "Self-Ask" [22] instructs LLMs to determine semantic gaps in a task and proactively ask themselves follow-up questions before returning a final answer. This mirrors our observation that when ChatGPT o3 poses clarifying questions, it builds deeper semantic context, leading to focused and accurate reasoning. Recent frameworks like Filter-Supervisor Self-Correction (FS-C) [23] highlight the importance of models examining multiple reasoning paths and self-verifying intermediate steps. In our case, ChatGPT 3.5's self-correction loop, where it detects and revises an incorrect rule mid-session, exemplifies this technique: engaging the model in multi-stage reflexive reasoning dramatically improved its rule derivation accuracy. Zero-Shot and Few-Shot Chain-of-Thought (CoT) prompting in LLMs [24] dramatically enhances performance on symbolic and mathematical reasoning tasks. This is seen in our structured query interactions with ChatGPT 3.5 and o3 improved rule induction and hypothesis validation.

## VII. CONCLUSION

This research presents the Primender sequence, a novel mathematical construct defined through a hybrid rule that blends classical primality with symbolic digit-based logic. The originality lies not only in the discovery of this sequence but also in its repurposing as a symbolic benchmark to evaluate the reasoning capabilities of Large Language Models (LLMs). Through the design of structured prompts and hypothesis testing, we demonstrate how models respond to abstract rules. Our proposed Symbolic Constructivist Evaluation Methodology (SCEM) offers a systematic and reproducible way to assess the depth of reasoning, making this work a valuable contribution at the intersection of number theory and artificial intelligence. This approach bridges mathematics, artificial intelligence, and software engineering, highlighting the role of symbolic reasoning in LLM interpretability and explainability.

The interdisciplinary relevance of this study lies in its ability to test, compare, and inspire LLM capabilities using interpretable symbolic structures. Our findings reveal that LLMs engaging in explanatory questioning outperform others in symbolic tasks - an insight aligned with emerging prompt-engineering research. Self-Ask and CoT prompting validate that models perform better when they autonomously question and expand prompt context. FS-C-style self-verification loops reflect observed benefits of internal reflection and correction in reasoning quality. However, looking at the significant error rates in all the models, it is clear that current state of artificial or generative intelligence is not sufficient for true symbolic inferences. The Primender benchmark is built around explicit symbolic inference – i.e. deducing a hidden generative rule from a sequence of symbols – and symbolic manipulation – applying or testing a rule on new elements. We found that even the LLM's chain of thought turns out to be a pattern extension and investigating the correctness of a proposed rule or hypothesis. However, they fail when tested to apply random algorithms or deducing logical rules on new sequences involving puzzle like pattern inference, in our case our newly discovered mathematical sequence - Primender.

Future work could include integrating such benchmarks into existing evaluation frameworks, exploring multimodal reasoning, expanding the Primender framework to new sequences, integrating with cognitive science evaluations, or developing new analysis tools for number theory including the study of primes. Overall, this research contributes a scalable, interpretable, and interdisciplinary framework to assess and enhance the cognitive depth of language models, advancing the frontier of human–AI symbolic collaboration.